\documentclass[11pt,a4paper]{article}
\usepackage[hyperref]{acl2019}
\usepackage{times}
\usepackage{latexsym}
\usepackage{booktabs}
\usepackage{csquotes}
\usepackage{graphicx}
\usepackage{natbib}
\usepackage{pbox}
\usepackage{url}

\newcommand{\humanperson}[0]{\includegraphics[width=.02\textwidth]{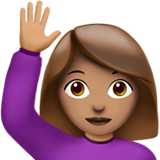}} 
\newcommand{\humanpersona}[0]{\includegraphics[width=.02\textwidth]{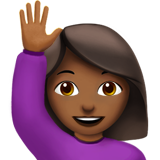}} 
\newcommand{\humanpersone}[0]{\includegraphics[width=.02\textwidth]{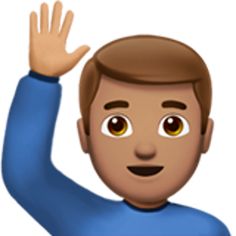}} 
\newcommand{\humanpersoni}[0]{\includegraphics[width=.02\textwidth]{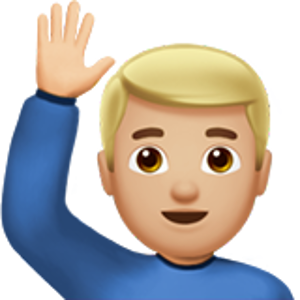}} 
\newcommand{\humanpersono}[0]{\includegraphics[width=.02\textwidth]{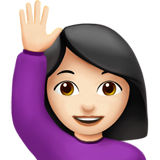}} 
\newcommand{\bert}[0]{\includegraphics[width=.02\textwidth]{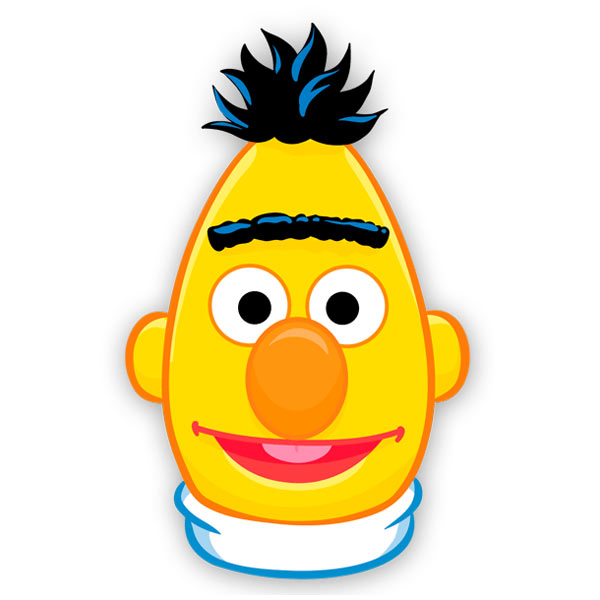}} 
\newcommand{\bigbird}[0]{\includegraphics[width=.02\textwidth]{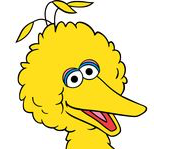}} 

\usepackage{xcolor}
\definecolor{mymaroon}{RGB}{153, 0, 0}
\definecolor{mygreen}{RGB}{96, 128, 0}


\aclfinalcopy 

\title{Human vs. Muppet: A Conservative Estimate of Human\\Performance on the GLUE Benchmark\\}

\author{
Nikita Nangia$^{1}$\\
\texttt{nikitanangia@nyu.edu}
\And
Samuel R.~Bowman$^{1,2,3}$\\
\texttt{bowman@nyu.edu}
\AND
$^{1}$\normalfont Center for Data Science\\New York University\And
$^{2}$\normalfont Dept. of Linguistics\\New York University\And
$^{3}$\normalfont Dept. of Computer Science\\New York University
} 

\date{}

\begin{document}
\maketitle

\begin{abstract}
The GLUE benchmark \citep{wang2018glue} is a suite of language understanding tasks which has seen dramatic progress in the past year, with average performance moving from 70.0 at launch to 83.9, state of the art at the time of writing (May 24, 2019). Here, we measure human performance on the benchmark, in order to learn whether significant headroom remains for further progress. We provide a conservative estimate of human performance on the benchmark through crowdsourcing: Our annotators are non-experts who must learn each task from a brief set of instructions and 20 examples. In spite of limited training, these annotators robustly outperform
the state of the art on six of the nine GLUE tasks and achieve an average score of 87.1. Given the fast pace of progress however, the headroom we observe is quite limited. To reproduce the data-poor setting that our annotators must learn in, we also train the BERT model \citep{devlin:2018} in limited-data regimes, and conclude that low-resource sentence classification remains a challenge for modern neural network approaches to text understanding.
\end{abstract}

\begin{table*}[t]\small
    \centering
    \setlength\tabcolsep{5pt}
    \begin{tabular}{lrrrrrrrrrr}
    \toprule
    & & \multicolumn{2}{c}{\bf{Single Sentence}} & \multicolumn{3}{c}{\bf{Sentence Similarity}} & \multicolumn{4}{c}{\bf{Natural Language Inference}} \\
     & \bf{Avg} & \bf{CoLA} & \bf{SST-2} & \bf{MRPC} & \bf{STS-B} & \bf{QQP} & \bf{MNLI} & \bf{QNLI} & \bf{RTE} & \bf{WNLI} \\
    \midrule
    \textit{Training Size} & - & \it8.5k & \it67k & \it3.7k & \it7k & \it364k & \it393k & \it108k & \it2.5k & \it634 \\
    \midrule
    Human \humanperson & \bf87.1 & \bf66.4 & \bf97.8 & 80.8/86.3 & \bf92.7/92.6 & 80.4/59.5 & \bf92.0/92.8 & 91.2 & \bf93.6 & \bf95.9 \\
    BERT \bert & 80.5 & 60.5 & 94.9 & 85.4/89.3 & 87.6/86.5 & \bf89.3/72.1 & 86.7/85.9 & \bf92.7 & 70.1 & 65.1 \\
    BigBird \bigbird & 83.9 & 65.4 & 95.6 & \bf88.2/91.1 & 89.5/89.0 & \bf89.6/72.7 & 87.9/87.4 & 95.8* & 85.1 & 65.1 \\
    
    \rule{0pt}{2.4ex} 
    $\Delta_{bert}$ (\humanperson\ -\bert) &  6.6 &  5.9 & 2.9 &  -4.6/-3.0 & 5.1/6.1 & -8.9/-12.6 & 5.3/6.9 & -1.5 & 23.5 & 30.8\\
    \ $\Delta_{bird}$ (\humanperson\ -\bigbird) & 3.2 & 1.0 &  2.2 & -7.4/-4.8 & 3.2/3.6 & -9.2/-13.2 & 4.1/5.4 & -4.6* & 8.5 & 30.8\\
    
    \midrule
    \midrule
    \multicolumn{11}{l}{\textit{Performance on subset with 5-way annotator agreement}} \\
    \midrule
    Human \humanperson\humanpersone\humanpersoni\humanpersona\humanpersono & 93.7 & 83.6 & 100.0 & 90.2/93.6 & 98.9/94.7 & 89.4/74.1 & 98.5/99.2 & 95.1 & 97.4 & 97.5 \\
    BERT \bert & 83.5 & 69.2 & 97.5 & 88.9/92.7 & 95.8/82.3 & 92.5/78.0 & 96.4/90.8 & 93.6 & 73.0 & 59.3 \\
    $\Delta$ (\humanperson\humanpersone\humanpersoni\humanpersona\humanpersono\ -\bert)& 10.2 & 14.4 & 2.5 & 1.3/0.9 & 3.1/12.4 & -3.1/-3.9 & 2.1/8.4 & 1.5 & 24.4 & 38.2 \\
    \midrule
    \midrule 
    \multicolumn{11}{l}{\textit{BERT fine-tuned on less data}} \\
    \midrule
    BERT-5000 & 75.8 & 57.6 & 92.0 & 85.4/89.3 & 87.1/85.8 & 82.2/61.0 & 76.4/76.9 & 89.2 & 69.2 & 65.1 \\
    BERT-1000 & 70.7 & 49.0 & 90.4 & 78.5/84.3 & 83.6/82.3 & 77.8/55.8 & 66.5/68.3 & 86.6 & 65.6 & 65.1 \\
    BERT-500 & 68.5 & 37.2 & 88.1 & 74.0/80.7 & 77.3/75.2 & 75.4/51.2 & 61.8/63.0 & 85.7 & 61.5 & 65.1 \\
    \bottomrule
    \end{tabular}
    \caption{GLUE test set results. The \textit{Human} baseline numbers are estimated using no more than 500 test examples. All the BERT scores 
    are for \textsc{BERT-large}. As in the original GLUE paper,  we report the Matthews correlation coefficient for CoLA. For MRPC and Quora, we report accuracy then F1. For STS-B, we report Pearson then Spearman correlation. 
    For MNLI, we report accuracy on the matched then mismatched test sets.  For all other tasks we report accuracy. The \textit{Avg} column shows the overall GLUE score: an average across each row, weighting each task equally. 
    The $\Delta_{bert}$ and $\Delta_{bird}$ rows show the difference between the \textit{Human} performance baseline and BERT and BigBird respectively. 
    The starred(*) numbers for BigBird on QNLI show performance on the new version of QNLI, while all other QNLI numbers are on the original version.
    The second section shows \textit{Human} and BERT performance on the subset of the test set where there is unanimous, 5-way annotator agreement, the $\Delta$ row is the difference between them.
    \textit{Training Size} gives the number of examples in the full training set for each task. The BERT-5000/1000/500 rows show test set results for BERT fine-tuned on no more than 5k, 1k, and 500 examples respectively. Though MRPC and RTE have fewer than 5k examples, we rerun BERT fine-tuning and report these results in the BERT-5000 row.} 
    \label{tab:main}
\end{table*}

\section{Introduction}
This past year has seen tremendous progress in building general purpose models that can learn good language representations across a range of tasks and domains \citep{mccann2017cove, matthew2018elmo, devlin:2018, howard2018, bigbird2019}. Reusable models like these can be readily adapted to different language understanding tasks and genres. The General Language Understanding Evaluation \citep[GLUE;][]{wang2018glue} benchmark is designed to evaluate such models. GLUE is built around nine sentence-level natural language understanding (NLU) tasks and datasets, including instances of natural language inference, sentiment analysis, acceptability judgment, sentence similarity, and common sense reasoning.

The recent BigBird model \citep{bigbird2019} ---a fine-tuned variant of the BERT model \citep{devlin:2018}---is state-of-the-art on GLUE at the time of writing, with the original BERT right at its heels. Both models perform impressively enough on GLUE to prompt some increasingly urgent questions: How much better are humans at these NLP tasks? Do standard benchmarks have enough headroom to meaningfully measure further progress? In the case of one prominent language understanding task with a known human performance number, SQuAD 2.0 \cite{rajpurkar2018}, models built on BERT come extremely close to human performance.\footnote{\url{https://rajpurkar.github.io/SQuAD-explorer/}} On the recent Situations With Adversarial Generations \citep[SWAG;][]{zellers2018} dataset, BERT \textit{outperforms} individual expert human annotators. 
In this work, we estimate human performance on the GLUE test set to determine which tasks see substantial remaining headroom between human and machine performance. 

While human performance or interannotator agreement numbers have been reported on some GLUE tasks, the data collection methods used to establish those baselines vary substantially. To maintain consistency in our reported baseline numbers, and to ensure that our results are at least roughly comparable to numbers for submitted machine learning models, we collect annotations using a uniform method for all nine tasks. 

We hire crowdworker annotators: For each of the nine tasks, we give the workers a brief training exercise on the task, ask them to annotate a random subset of the test data, and then collect \textit{majority vote} labels from five annotators for each example in the subset. Comparing these labels with the ground-truth test labels yields an overall GLUE score of 87.1---well above BERT's 80.5 and BigBird's 82.9---and yields single-task scores that are substantially better than both on six of nine tasks. However, in light of the pace of recent progress made on GLUE, the gap in most tasks is relatively small. The one striking exception is the data-poor Winograd Schema NLI Corpus \citep[WNLI; based on][]{Levesque:2012}, in which humans outperform machines by over 30 percentage points.

To reproduce the data-poor training regime of our annotators, and of WNLI, we investigate BERT's performance on data-poor versions of the other GLUE tasks and find that it suffers considerably in these low-resource settings. Ultimately however, BERT's performance seems genuinely close to human performance and leaves limited headroom in GLUE.

\section{Background and Related Work}

\paragraph{GLUE}
GLUE \citep{wang2018glue} is composed of nine sentence or sentence-pair classification or regression tasks: MultiNLI \citep{williams2018mnli}, RTE (competition releases 1--3 and 5, merged and treated as a single binary classification task; \citealt{rte1}, \citealt{rte2}, \citealt{rte3}, \citealt{rte5}), QNLI (an answer sentence selection task based on SQuAD; \citealt{rajpurkar2016}),\footnote{Our human performance numbers for QNLI are on the original test set since we collected data before the release of the slightly revised second test set. \textsc{BERT-large}'s performance went up by 1.6 percentage points on the new test set, and \textsc{BERT-base}'s performance saw a 0.5 point increase. This suggests that our human performance number represents a reasonable---if very conservative---approximation of human performance on QNLI.} 
and WNLI test natural language inference. WNLI is derived from private data created for the Winograd Schema Challenge \citep{Levesque:2012}, which specifically tests for common sense reasoning.
The Microsoft Research Paraphrase Corpus \citep[MRPC;][]{dolan2005}, the Semantic Textual Similarity Benchmark (STS-B; \citeauthor{cer2017sts}, \citeyear{cer2017sts}), and Quora Question Pairs (QQP)\footnote{\url{https://data.quora.com/First-Quora-Dataset-Release-Question-Pairs}}
test paraphrase and sentence similarity evaluation.  The Corpus of Linguistic Acceptability \citep[CoLA;][]{warstadt2018} tests grammatical acceptability judgment. Finally, the Stanford Sentiment Treebank \citep[SST;][]{socher2013sst} tests sentiment analysis.

\paragraph{Human Evaluations on GLUE Tasks}
\citet{warstadt2018} report human performance numbers on CoLA as well. Using the majority decision from five expert annotators on 200 examples, they get a Matthews correlation coefficient (MCC) of 71.3.
\citet{Bender2015establishing} also estimates human performance on the original public Winograd Schema Challenge (WSC) data. They use crowdworkers and report an average accuracy of 92.1\%. The RTE corpus papers report inter-annotator agreement numbers on their test sets: 80\% on RTE-1, 89.2\% on RTE-2, 87.8\% on RTE-3, and 97.02\% on RTE-5.
\citet{wang2018glue} report human performance numbers on GLUE's manually curated diagnostic test set. The examples in this test set are natural language inference sentence pairs that are tagged for a set of linguistic phenomena. They use expert annotators and report an average $R_3$ coefficient of 0.8.

\section{Data Collection Method}\label{sec:data}

To establish human performance on GLUE, we hire annotators through the 
Hybrid\footnote{\url{http://www.gethybrid.io}} data collection platform, which is similar to Amazon's Mechanical Turk. Each worker first completes a short training procedure then moves on to the main annotation task. For the annotation phase, we tune the pay rate for each task, with an average rate of \$17/hr. The training phase has a lower, standard pay rate, with an average pay of \$7.6/hr.

\paragraph{Training} In the training phase for each GLUE task, each worker answers 20 random examples from the task development set. Each training page links to instructions that are tailored to the task, and shows five examples.\footnote{All the task specific instructions and FAQ pages used can be found at \url{https://nyu-mll.github.io/GLUE-human-performance/}} The answers to these examples can be revealed by clicking on a ``Show" button at the bottom of the page. We ask the workers to label each set of examples and check their work so they can familiarize themselves with the task. Workers who get less than 65\% of the examples correct during training do not qualify for the main task. This is an intentionally low threshold meant only to encourage a reasonable effort. Our platform cannot fully prevent workers from changing their answers after viewing the correct labels, so we cannot use the training phase as a substantial filter. (See Appendix~\ref{app:train} for details on the training phase.)

\paragraph{Annotation} We randomly sample 500 examples from each task's test set for annotation, with the exception of WNLI where we sample 145 of the 147 available test examples (the two missing examples are the result of a data preparation error). For each of these sampled data points, we collect five annotations from five different workers (see Appendix~\ref{app:test}). 
We use the test set since the test and development sets are qualitatively different for some tasks, and we wish to compare our results directly with those on the GLUE leaderboard.

\section{Results and Discussions}

To calculate the human performance baseline, we take the majority vote across the five crowd-sourced annotations. In the case of MultiNLI, since there are three possible labels---\textit{entailment}, \textit{neutral}, and \textit{contradiction}---about 2\% of examples see a tie between two labels. For these ties, we take the label that is more frequent in the development set. In the case of STS-B, we take an average of the scalar annotator labels. Since we only collect annotations for a subset of the data, we cannot access the test set through the GLUE leaderboard interface, we instead submit our predictions to the GLUE organizers privately.

We compare human performance to BERT and BigBird. The human performance numbers in Table~\ref{tab:main} shows that overall our annotators \textit{stick it to the Muppets} on GLUE. However on MRPC, QQP, and QNLI, Bigbird and BERT \textit{outperform} our annotators. The results on QQP are particularly surprising: BERT and BigBird score over 12 F1 points better than our annotators. Our annotators, however, are only given 20 examples and a short set of instructions for training, while BERT and BigBird are fine-tuned on the 364k-example QQP training set. 
In addition, we find it difficult to compose concise instructions for QQP that actually match the supplied labels. We do not have access to the material used to create the dataset, and we find it difficult to infer simple instructions from the data (sample provided in Appendix~\ref{sec:exs}). If given more training data, it is possible that our annotators could better learn relatively subtle label definitions that better fit the corpus.


\paragraph{Unanimous Vote} To investigate the possible effect of ambiguous label definitions, we look at human performance when there is 5-way annotator agreement. Using unanimous agreement, rather than majority agreement, has the potential effect of filtering out examples of two kinds: those for which our supplied annotation guidelines don't provide clear advice and those for which humans understand the expectations of the task but find the example genuinely difficult or uncertain. 
To disentangle the two effects, we also look at BERT results on this subset of the test set, as BERT's use of large training sets means that it should only suffer in the latter cases. We get consent from the authors of BERT to work in cooperation with the GLUE team to measure BERT's performance on this subset, which we show in  Table~\ref{tab:main}. Overall, we see the gap widen between the human baseline and BERT by 3.1 points. The largest shifts in performance are on CoLA, MRPC, QQP, and WNLI. The relative jumps in performance on MRPC and QQP 
support the claim that human performance is hurt by imprecise guidelines and that the use of substantially more training data gives BERT an edge on our annotators.

In general, BERT needs large datasets to fine-tune on. This is further evidenced by its performance discrepancy between MultiNLI and RTE: human performance is similar for the two, whereas BERT shows a 16.2 percentage point gap between the two datasets.
Both MultiNLI and RTE are textual entailment datasets, but MultiNLI's training set is quite large at 393k examples, while the GLUE version of RTE has only 2.5k examples. However, BigBird does not show as large a gap, which may be because it employs a multi-task learning approach which fine-tunes the model for all sentence-pair tasks jointly. Their RTE classifier, for example, benefits from the large training dataset for the closely related MultiNLI task.

\paragraph{Low-Resource BERT Baseline} To understand the impact of abundant target tasks on the limited headroom that we observe, we train several additional baselines. 
In these, we fine-tune BERT on 5k, 1k, and 500 examples for each GLUE task (or fewer for tasks with fewer training examples).
We use BERT for this analysis because the authors have released their code and have provided pretrained weights for the model. We use their publicly available implementation of \textsc{BERT-large}, their pretrained weights as the initialization for fine-tuning on the GLUE tasks, and the hyperparameters they report.
We see a precipitous drop in performance on most tasks with large datasets, with the exception of QNLI. A possible partial explanation is that both QNLI and the BERT training data come from English Wikipedia.
On MRPC and QQP however, BERT's performance drops below human performance in the 1k- and 500-example settings. 
On the whole, we find that BERT suffers in low-resource settings. These results are in agreement with the findings in \citet{phang2018sentence} who conduct essentially the same experiment.

\paragraph{CoLA} Our human performance number on CoLA is 4.9 points below what was reported in \citet{warstadt2018}.  We believe this discrepancy is because they use linguistics PhD students as expert annotators while we use crowdworkers. This further supports our belief that our human performance baseline is a conservative estimate, and that higher performance is possible, particularly with more training.

\paragraph{WNLI} No system on the GLUE leaderboard has managed to exceed the performance of the most-frequent-class baseline on WNLI, and several papers that propose methods for GLUE justify their poor performance by asserting that the task must be somehow broken.\footnote{\citet{devlin:2018}, for example, mention that they avoid ``the problematic WNLI set''.} 
WNLI's source Winograd Schema data was constructed so as not to include any statistical cues that a simple machine learning system can exploit, which can make it quite difficult. The WNLI test set shows one of the \textit{highest} human performance scores of the nine GLUE tasks, reflecting its status as a corpus constructed and vetted by artificial intelligence experts. This affirms that tasks like WNLI with small training sets (634 sentence pairs) and no simple cues remain a serious (and sometimes unacknowledged) blind spot for modern neural network sentence understanding methods.

\section{Conclusion} 
This paper presents a conservative estimate of human performance to serve as a target for the GLUE sentence understanding benchmark. We obtain this baseline with the help of crowdworker annotators. We find that state-of-the-art models like BERT are not far behind human performance on most GLUE tasks. But we also note that, when trained in low-resource settings, BERT's performance falls considerably. Given these results, and the continued difficulty neural methods have with the Winograd Schema Challenge, we argue that future work on GLUE-style sentence understanding tasks might benefit from a focus on learning from smaller training sets. In work subsequent to the main results of this paper, we have prepared such a benchmark in the GLUE follow-up SuperGLUE \citep{superglue2019}.

\section*{Acknowledgments}
This work was made possible in part by a donation to NYU from Eric and Wendy Schmidt made by recommendation of the Schmidt Futures program and by funding from Samsung Research. We gratefully acknowledge the support of NVIDIA Corporation with the donation of a Titan V GPU used at NYU for this research. We thank Alex Wang and Amanpreet Singh for their help with conducting GLUE evaluations, and we thank Jason Phang for his help with training the BERT model.

\bibliography{acl2019}
\bibliographystyle{acl_natbib}
 
\appendix

\section{Crowd-Sourced Data Collection}
\subsection{Training Phase}\label{app:train}
During training, we provide a link to task-specific instructions. As an example, the instructions for CoLA are shown in Table~\ref{tab:cola}. The instructions for each task follows the same format: briefly describing the annotator's job, explaining the labels, and providing at least one example.

In addition to the task-specific instructions, we provide general instructions about the training phase. An example is given in Table~\ref{tab:traininst}. Lastly, we provide a link to an FAQ page. The FAQ page addresses the balance of the data. If the labels are balanced, we tell the annotators so. If the labels are not balanced, we assure the annotators that they need not worry about assigning one label more frequently. For most tasks we also describe where the data comes from, e.g. news articles. All of the task specific instructions and FAQ pages can be found at \url{nyu-mll.github.io/GLUE-human-performance/}.

On each training page, each annotator is given five examples to annotate. At the bottom of the page, there is a ``Show" button which reveals the ground truth labels. If their submitted answer is incorrect, the label is shown in red, otherwise it is shown in black. In the instructions, the annotator is asked to check their work with this button. Given this procedure, we cannot prevent the annotators from changing their answer after viewing the ground truth labels.

\begin{table}[!t]
\begin{tabular}{p{0.5\textwidth}}
\toprule
\vspace{0.2cm}

New York University's Center for Data Science is collecting your answers for use in research on computer understanding of English. Thank you for your help!\\

We will present you with a sentence someone spoke. \textcolor{mymaroon}{Your job is to figure out, based on this sentence, if the speaker is a native speaker of English. You should ignore the general topic of the sentence and focus on the fluency of the sentence}.

\begin{itemize}
\item Choose correct if you think the sentence sounds fluent and you think it was spoken by a native-English speaker. Examples:
    \begin{itemize}\vspace{-0.1cm}
        \item[] \textcolor{mygreen}{\textit{``A hundred men surrounded the fort.”}}
        \item[] \textcolor{mygreen}{\textit{``Everybody who attended last week’s huge rally, whoever they were, signed the petition."}}
        \item[] \textcolor{mygreen}{\textit{``Where did you go and who ate what?"}}
    \end{itemize}

\item Choose incorrect if you think the sentence does not sound completely fluent and may have been spoken by a non-native English speaker. Examples:\vspace{-0.1cm}
    \begin{itemize}
        \item[] \textcolor{mygreen}{\textit{``Sue gave to Bill a book.”}}
        \item[] \textcolor{mygreen}{\textit{``Mary came to be introduced by the bartender and I also came to be."}}
        \item[] \textcolor{mygreen}{\textit{``The problem perceives easily."}}
    \end{itemize}
\end{itemize}
More questions? See the FAQ page.\medskip\\

\bottomrule
\end{tabular}

\caption{The instructions given to crowd-sourced worker for the CoLA task. While the instructions were tailored for each task in GLUE, they all followed a similar format.}
\label{tab:cola}
\end{table}

\begin{table*}[!t]
\begin{tabular}{p{0.95\textwidth}}
\toprule
\vspace{0.2cm}

This project is a training task that needs to be completed before working on the main project on Hybrid named \textcolor{red}{Human Performance: CoLA}. For this CoLA task, we have the true label and we want to get information on how well people do on the task. This training is short but is designed to help you get a sense of the questions and the expected labels.\\

Please note that the pay per HIT for this training task is also lower than it is for the main project Human Performance: CoLA. Once you are done with the training, please proceed to the main task!\\

In this training, you must answer all the questions on the page and then, to see how you did, click the Show button at the bottom of the page before moving onto the next HIT. The Show button will reveal the true labels. If you answered correctly, the revealed label will be in black, otherwise it will be in red. Please use this training and the provided answers to build an understanding of what the answers to these questions looks like (the main project, Human Performance: CoLA, does not have the answers on the page).\medskip\\

\bottomrule
\end{tabular}

\caption{Instructions about the training phase provided to workers. This example is for CoLA training. The only change in instructions for other tasks is the name of the task.}
\label{tab:traininst}
\end{table*}

\subsection{Annotation Phase}\label{app:test}
In the main data collection phase we provide the annotators with a link to the same task-specific instructions (Table~\ref{tab:cola}) and FAQ page used during the training phase. We enforce the training phase as a qualification for annotation, so crowdworkers cannot participate in annotation without first completing the associated training.

\section{QQP Example}\label{sec:exs}
The 25 randomly sampled examples from the QQP development set are given in Tables~\ref{tab:qqp}, \ref{tab:qqp2}, and \ref{tab:qqp3}.

\begin{table*}[!t]
    \centering
    \begin{tabular}{l l r}
    \toprule
         \textbf{Question 1} & \textbf{Question 2} & \textbf{Label} \\
         \midrule\\
         
         \pbox{6.5cm}{\it What are the best resources for learning Ukrainian?} & \pbox{7cm}{\it What are the best resources for learning Turkish?} & 0 \\\\
         \midrule \\
         
         \pbox{7cm}{\it How much time will it take to charge a 10,000 mAh power bank?} & \pbox{7cm}{\it How much time does it takes to charge the power bank 13000mAh for full charge?} & 0 \\\\
         \midrule \\
         
         \pbox{7cm}{\it How do you know if you're in love?} & \pbox{7cm}{\it How can you know if you're in love or just attracted to someone?} & 1 \\\\
         \midrule \\
         
         \pbox{7cm}{\it Which are the best and affordable resorts in Goa?} & \pbox{7cm}{\it What are some affordable and safe beach resorts in Goa?} & 1 \\\\
         \midrule \\
         
         \pbox{7cm}{\it How winning money from YouTube?} & \pbox{7cm}{\it How do I make money from a YouTube channel?} & 1 \\\\
         
    \bottomrule
    \end{tabular}
    \caption{Five randomly sampled examples from QQP's development set. Pairs of sentences with a label of 1 are marked as paraphrases in QQP.}
    \label{tab:qqp}
\end{table*}

\begin{table*}
    \centering
    \begin{tabular}{l l r}
    \toprule
         \textbf{Question 1} & \textbf{Question 2} & \textbf{Label} \\
         \midrule\\
         
         \pbox{7cm}{\it What is actual meaning of life? Indeen, it depend on perception of people or other thing?} & \pbox{7cm}{\it What is the meaning of my life?} & 1 \\\\ 
         \midrule \\
         
         \pbox{7cm}{\it What is the difference between CC and 2S classes of travel in Jan Shatabdi express?} & \pbox{7cm}{\it What is TQWL in IRCTC wait list?} & 0 \\\\ 
         \midrule \\
            
         \pbox{7cm}{\it What would have happened if Hitler hadn't declared war on the United States after Pearl Harbor? } & \pbox{7cm}{\it What would have happened if the United States split in two after the revolutionary war?} & 0 \\\\ 
         \midrule \\
         
         \pbox{7cm}{\it Will it be a problem if a friend deposits 4 lakhs in my savings bank account and I don't have a source of income to show?} & \pbox{7cm}{\it I am 25.5 year old boy with a B.Com in a sales job having a package of 4 LPA. I will be married in less than a year. I want to quit my job and start my own business with the savings I have of 2 Lakh. Is this an ideal situation to take a risk?} & 0 \\\\ 
         \midrule \\
         
         \pbox{7cm}{\it What should you do if you meet an alien?} & \pbox{7cm}{\it What could be the possible conversation between humans and aliens on their first meeting?} & 0 \\\\ 
         \midrule \\
         
         \pbox{7cm}{\it Why can't I ask any questions on Quora?} & \pbox{7cm}{\it Can you ask any question on Quora?} & 0 \\\\ 
         \midrule \\
         
         \pbox{7cm}{\it Should I move from the USA to India?} &  \pbox{7cm}{\it Moving from usA to India?} & 1 \\\\ 
         \midrule \\
         
         \pbox{7cm}{\it Which European countries provide mostly free university education to Indian citizen?} & \pbox{7cm}{\it What countries provide free education to Indian students?} & 0 \\\\
         \midrule \\
         
         \pbox{7cm}{\it I got 112 rank in CDAC (A+B+C). My subject of interest is VLSI. Is there any chance that I would get CDAC Pune, Noida for VLSI?} & \pbox{7cm}{Suggest some good indian youtube channels for studying Aptitude?} & 0 \\\\
         \midrule \\
         
         \pbox{6cm}{\it What are the positives and negatives of restorative justice?} & \pbox{7cm}{\it Is Vengence and Justice opposite?} & 0 \\\\
         
    \bottomrule
    \end{tabular}
    \caption{Another ten randomly sampled examples from QQP's development set. Pairs of sentences with a label of 1 are marked as paraphrases in QQP.}
    \label{tab:qqp2}
\end{table*}

\begin{table*}
    \centering
    \begin{tabular}{l l r}
    \toprule
         \textbf{Question 1} & \textbf{Question 2} & \textbf{Label} \\
         \midrule\\
         
         \pbox{7cm}{\it What's a good way to make money through effort?} & \pbox{7cm}{\it How do I make money without much effort?} & 0 \\\\
         \midrule \\
         
         \pbox{7cm}{\it What is the meaning of life? Whats our purpose on Earth?} & \pbox{7cm}{\it What actually is the purpose of life?} & 1 \\\\
         \midrule \\
         
         \pbox{7cm}{\it Which among five seasons (summer, winter, autumn, spring, rainy) is most favourable for farming and cultivating of crops?} & \pbox{7cm}{\it Which among the five seasons (summer, winter, rainy, spring, autumn) is better for farming and cultivating of crops?} & 1 \\\\
         \midrule \\
         
         \pbox{6.5cm}{\it How can I find the real true purpose of my life?} & \pbox{7cm}{\it What should one do to find purpose of one's life?} & 1 \\\\
         \midrule \\
         
         \pbox{7cm}{\it Is Donald Trump likely to win the 2016 election (late 2015 / early 2016)?} & \pbox{7cm}{\it What will Donald Trump's response be if he doesn't win the 2016 presidential election?} & 0 \\\\
         \midrule \\
         
         \pbox{7cm}{\it What is the easiest and cheapest way to lose weight fast?} & \pbox{7cm}{\it What are the easiest and the fastest ways to lose weight?} & 1 \\\\
         \midrule \\
         
         \pbox{7cm}{\it Why are basically all of my questions on Quora marked as 'needing improvement'? Am I that bad?} & \pbox{7cm}{\it Why do questions get marked for 'needing improvment' when they clearly don't?} & 1 \\\\
         \midrule \\
         
         \pbox{7cm}{\it What are some of the most visually stunning apps?} & \pbox{7cm}{\it What are the most visually stunning foods?} & 0 \\\\
         \midrule \\ 
         
         \pbox{7cm}{\it What are some of the good hotels near chennai central railway station?} & \pbox{7cm}{\it Best places to eat in Chennai?} & 0 \\\\
         \midrule \\
         
         \pbox{7cm}{\it How do you prepare for a job interview?} & \pbox{7cm}{\it How do I prepare for my first job interview?} & 1 \\\\
    \bottomrule
    \end{tabular}
    \caption{Another ten randomly sampled examples from QQP's development set. Pairs of sentences with a label of 1 are marked as paraphrases in QQP.}
    \label{tab:qqp3}
\end{table*}

\end{document}